\documentclass[lettersize,journal]{IEEEtran}
\usepackage{amsmath,amsfonts}
\usepackage{algorithmic}
\usepackage{algorithm}
\usepackage{array}
\usepackage{textcomp}
\usepackage{stfloats}
\usepackage{url}
\usepackage{verbatim}
\usepackage{graphicx}
\usepackage{cite}
\usepackage{xcolor}
\usepackage{subfigure}
\begin{document}

% \title{Multimodal Channel Estimation via \\ Flow Matching with Diffusion Transformers}
\title{Improving Channel Estimation via \\ Multimodal Diffusion Models with Flow Matching}

\author{Xiaotian Fan,~\IEEEmembership{Graduate Student Member,~IEEE}, Xingyu Zhou,~\IEEEmembership{Graduate Student Member,~IEEE}, \\Le Liang,~\IEEEmembership{Member,~IEEE}, Xiao Li,~\IEEEmembership{Member,~IEEE}, and Shi Jin,~\IEEEmembership{Fellow,~IEEE}
        % <-this % stops a space
\thanks{X. Fan, X. Zhou, L. Liang, Xiao Li and S. Jin are with the School of Information Science and Engineering, Southeast University, Nanjing 210096, China (e-mail: xt\_fan@seu.edu.cn; xy\_zhou@seu.edu.cn; lliang@seu.edu.cn; li\_xiao@seu.edu.cn; jinshi@seu.edu.cn). L. Liang is also with Purple Mountain Laboratories, Nanjing 211111, China.}}% <-this % stops a space


% % column for its text to clear the IEEEpubid mark.

\maketitle

\begin{abstract}
Deep generative models offer a powerful alternative to conventional channel estimation by learning complex channel distributions. By integrating the rich environmental information available in modern sensing-aided networks, this paper proposes MultiCE-Flow, a multimodal channel estimation framework based on flow matching and diffusion transformer (DiT). We design a specialized multimodal perception module that fuses LiDAR, camera, and location data into a semantic condition, while treating sparse pilots as a structural condition. These conditions guide a DiT backbone to reconstruct high-fidelity channels. Unlike standard diffusion models, we employ flow matching to learn a linear trajectory from noise to data, enabling efficient one-step sampling. By leveraging environmental semantics, our method mitigates the ill-posed nature of estimation with sparse pilots. Extensive experiments demonstrate that MultiCE-Flow consistently outperforms traditional baselines and existing generative models. Notably, it exhibits superior robustness to out-of-distribution scenarios and varying pilot densities, making it suitable for environment-aware communication systems.
\end{abstract}


\section{Introduction}

\IEEEPARstart{M}{assive} multiple-input multiple-output (MIMO) systems % 这里说MIMO够吗，需要MIMO-OFDM和后文对应吗
rely critically on accurate channel state information (CSI) to achieve their promised gains in spectral efficiency~\cite{OVERMIMO}. However, acquiring accurate CSI through pilot transmissions incurs significant overhead, posing a major challenge for practical deployment. Under pilot-constrained scenarios, traditional least squares (LS) estimation performs poorly in low signal-to-noise ratio (SNR) environments. While linear minimum mean square error (LMMSE) estimation, though offering better performance, introduces high computational complexity due to the need for accurate channel covariance matrices. Although methods exploiting channel sparsity in the angular or delay domain can reduce pilot overhead, these approaches struggle to capture the complicated characteristics of realistic channels, as the sparsity assumption is often an oversimplification of the true channel distribution.

% 主要要补一些比较新的文献引用
% 目前想到的一个是变分score的那篇
To address this limitation, recent research has shifted toward data-driven approaches that can model high-dimensional distributions directly from data \cite{AI}.
Among these, deep generative models have emerged as powerful tools. Diffusion models (DMs), in particular, have been widely adopted for wireless channel estimation due to their ability to learn complex priors.
Early works such as \cite{SGM} utilized score-based models to exploit learned channel priors. 
Addressing data scarcity and robust inference, a prior-aided estimation framework was developed to recover high-dimensional channels from noisy or quantized measurements via posterior sampling \cite{HGM}. Variational inference approaches further leveraged pre-trained diffusion priors to infer the channel posterior \cite{SGMVI}. To reduce computational costs, studies like \cite{LC,LCLDM} proposed lightweight diffusion architectures, enabling faster inference for real-time applications.

However, most diffusion-based estimators rely solely on pilot-to-channel mapping, largely ignoring the physical propagation environment. While recent works have begun to explore the connection between the environment and wireless channels \cite{PF, MME}, further progress necessitates a shift toward more robust frameworks: deeply integrating multimodal environmental semantics and adopting foundation model architectures.

On the one hand, wireless channels are fundamentally determined by physical propagation. Multimodal sensors (LiDAR, cameras) and positioning information capture the environmental geometry that shapes these channels. By fusing these modalities, we can extract semantic priors that guide channel reconstruction, enabling accurate inference even with extremely sparse pilots.

On the other hand, diffusion models have evolved from U-Nets to transformer-based foundation models such as PixArt-$\alpha$ and stable diffusion 3 \cite{PixArt, SD3}, which achieve high-fidelity generation via conditioning on text prompts. Inspired by this, we formulate channel estimation as a conditional generation task: sparse pilots provide structural guidance, while multimodal data acts as the semantic prompt for reconstruction. To support this, we employ flow matching \cite{FM}, which learns direct transport paths from noise to data, enabling faster and higher-quality inference.

Based on this, we propose a novel multimodal channel estimation framework via flow matching with diffusion transformers, named MultiCE-Flow. We design a scalable diffusion transformer (DiT) to fuse heterogeneous modalities with sparse pilots. By learning velocity fields that transport noise directly to the channel distribution, our method enables fast, one-step estimation. Experiments show that the proposed method outperforms traditional and generative learning baselines while exhibiting superior robustness to out-of-distribution (OOD) scenarios.

% Based on these observations, we propose MultiCE-Flow, a flow matching-based diffusion transformer for multimodal channel estimation. We design a scalable DiT architecture to fuse heterogeneous modalities (LiDAR, camera, and location) with sparse pilot observations. The Transformer backbone handles variable-length inputs and scales to large multimodal datasets through self-attention. We formulate channel estimation as learning a velocity field that transports Gaussian noise directly to the channel distribution. Unlike iterative diffusion models, flow matching produces straighter trajectories that enable accurate one-step sampling, reducing inference time by orders of magnitude. To handle distribution shifts, we incorporate classifier-free guidance, allowing the model to dynamically balance between learned priors and environmental observations. Experiments show that our method outperforms traditional and generative baselines across diverse scenarios. More importantly, it demonstrates strong robustness to out-of-distribution environments, making it suitable for real-world 6G deployment where channel conditions vary widely.

\section{System Model and Multimodal Data}
% 小标题需要重新修改一下可能
\subsection{System Model}
We consider a downlink vehicle-to-infrastructure (V2I) massive MIMO-orthogonal frequency division multiplexing (OFDM) system. % 这样的写法看起来怪怪的
% operating at a carrier frequency of $f_c = 28$ GHz. 
The road side unit (RSU), acting as the base station, is equipped with a uniform linear array (ULA) of $N_{\mathrm{t}}$ transmit antennas, while each connected and autonomous vehicle (CAV) is equipped with a ULA of $N_{\mathrm{r}}$ receive antennas.
The system utilizes $N_{\mathrm{c}}$ subcarriers and a subcarrier spacing of $\Delta f$. The frequency-domain channel matrix $\mathbf{H}[k] \in \mathbb{C}^{N_{\mathrm{r}} \times N_{\mathrm{t}}}$ at the $k$-th subcarrier ($k = 0, \dots, N_{\mathrm{c}}-1$) is modeled as a superposition of $L$ propagation paths:
% derived from ray-tracing simulations
\begin{equation}
    \mathbf{H}[k] = \sum_{l=1}^{L} \mathbf{A}_l e^{-j 2\pi k \Delta f \tau_l},
\end{equation}
where $L$ denotes the number of multipath components. For the $l$-th path, $\mathbf{A}_l \in \mathbb{C}^{N_{\mathrm{r}} \times N_{\mathrm{t}}}$ represents the complex path gain matrix, and $\tau_l$ represents the propagation delay.
% where $L$ denotes the number of multipath components, including the Line-of-Sight (LOS) path and first-order reflections. For the $l$-th path, $\mathbf{A}_l \in \mathbb{C}^{N_r \times N_t}$ represents the complex path gain matrix incorporating antenna patterns and specific material properties (based on ITU-R P.527-5), and $\tau_l$ represents the propagation delay.

During the pilot transmission phase, the RSU transmits pilot symbols on a subset of subcarriers $\mathcal{P} \subset \{0, 1, \dots, N_{\mathrm{c}}-1\}$. To ensure orthogonality among transmit antennas, frequency-domain interleaved pilots are employed. Specifically, the pilot subcarriers are assigned to different transmit antennas in a cyclic manner, such that at any given pilot subcarrier $k \in \mathcal{P}$, only one transmit antenna $\mathrm{t}_k$ is active. The received signal at the CAV is given by:
\begin{equation}
    \mathbf{y}[k] = \mathbf{H}[k] \mathbf{x}[k] + \mathbf{n}[k] = \mathbf{h}_{\mathrm{t}_k}[k] s[k] + \mathbf{n}[k], \quad k \in \mathcal{P},
\end{equation}
where $\mathbf{x}[k] \in \mathbb{C}^{N_{\mathrm{t}} \times 1}$ is the sparse pilot vector with only one non-zero entry $s[k]$ corresponding to the active antenna $\mathrm{t}_k$,  $\mathbf{h}_{t_k}[k] \in \mathbb{C}^{N_{\mathrm{r}} \times 1}$ denotes the $\mathrm{t}_k$-th column of the channel matrix $\mathbf{H}[k]$, and $\mathbf{n}[k] \sim \mathcal{CN}(\mathbf{0}, \sigma^2 \mathbf{I}_{N_{\mathrm{r}}})$ is the additive white Gaussian noise (AWGN). The goal is to estimate the complete channel tensor $\mathbf{H} \in \mathbb{C}^{N_{\mathrm{r}} \times N_{\mathrm{t}} \times N_{\mathrm{c}}}$ across all subcarriers.

\subsection{Multimodal Data and Alignment}
% 小标题是否需要重新修改一下
We employ the Multimodal-Wireless dataset \cite{MultimodalWireless}, a large-scale dataset for sensing and communication. 
% using the CARLA simulator for physical environment modeling and the Sionna ray-tracer for wireless propagation. 
The dataset comprises 16 distinct scenarios featuring diverse urban layouts and traffic densities. 
% We partition 15 scenarios for training, validation and testing, reserving one complete scenario (Town07 single-lane road) as an OOD testing set to evaluate generalization capability.
Each scenario captures synchronized streams of communication and sensory data for CAVs navigating through the environment. The detailed specifications are as follows:

\begin{itemize}
    \item LiDAR: Each CAV is equipped with a 64-channel LiDAR sensor with a capturing range of 120 meters and a vertical field of view (FOV) from $-25^{\circ}$ to $2^{\circ}$. It generates approximately $30,000$ points per frame, providing a 3D geometric representation of the surrounding scatterers.
    \item Camera: To capture visual semantics, each CAV is equipped with four RGB cameras mounted to cover a full $360^{\circ}$ horizontal FOV (front, back, left, right). Each camera provides images with a resolution of $640 \times 480$.
    \item Positioning Information: The global coordinates and heading angles of both CAVs and RSUs are recorded.
\end{itemize}

In the constructed dataset, the wireless channel is sampled every $10$ ms while multimodal sensors operate at $100$ ms intervals, leading to a sampling rate mismatch. To address this, we use a one-to-many temporal alignment strategy: each sensor measurement at time $t$ is paired with $10$ consecutive channel samples from $t$ to $t+90$ ms. This is justified by the fact that the physical environment changes slowly compared to the $100$ ms sensor interval, allowing a single environmental observation to provide valid context for multiple channel estimation within that window.

For spatial alignment, we transform all spatial modalities into a unified CAV-centric coordinate frame to facilitate robust geometric feature learning. Specifically, we map raw global positions into the CAV's body frame, where the origin is placed at the CAV's center and the x-axis aligns with its heading direction. 
In this frame, the RSU position is represented as a relative vector rather than absolute global coordinates. This transformation eliminates biases from arbitrary coordinate origins and enables the model to focus on the relative spatial topology that directly governs wireless propagation.

\section{Proposed Method}
We propose MultiCE-Flow, a framework that reconstructs high-fidelity wireless channels from sparse pilots combined with rich environmental semantics.

\subsection{Overall Framework}
\begin{figure*}[!t]
\centering
\includegraphics[width=5.7in]{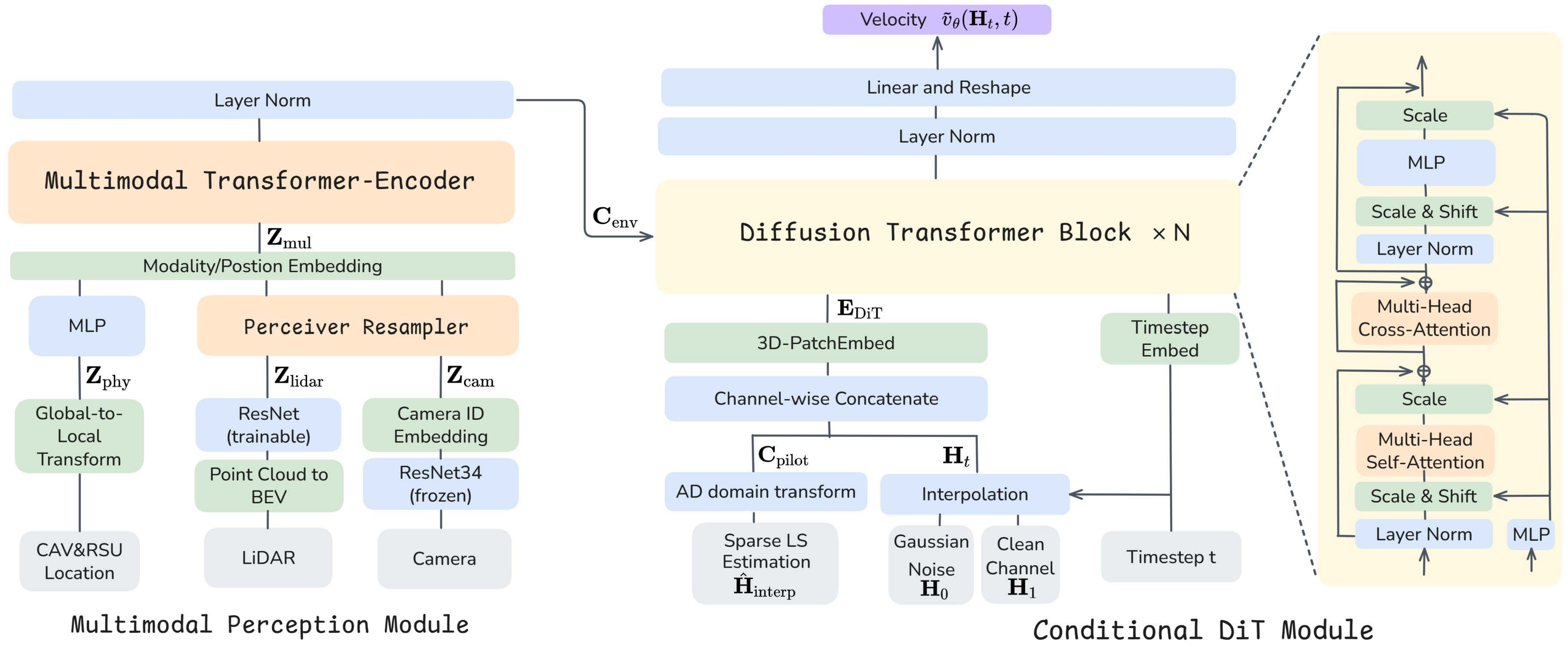}
\caption{The architecture of MultiCE-Flow. (Left) The multimodal perception module encodes LiDAR, camera, and positioning information into a fused semantic condition $\mathbf{C}_\mathrm{env}$. (Right) The DiT backbone uses flow matching to transport noise to the channel distribution, conditioned on both the environmental semantics and the pilot-based structural condition $\mathbf{C}_\mathrm{pilot}$.}
\label{fig_0}
\end{figure*}

MultiCE-Flow treats channel estimation as a conditional generation task. As shown in Fig.~\ref{fig_0}, the system comprises two core modules: a multimodal perception module and a conditional DiT. The workflow proceeds as follows:

Firstly, we extract a semantic condition from heterogeneous sensing data. LiDAR point clouds, camera images, and positioning information are processed to capture environmental context. These inputs are fused into a latent sequence $\mathbf{C}_{\mathrm{env}}$, acting as a semantic condition to refine the estimation. In parallel, we derive a structural condition from sparse pilot observations $\{\mathbf{Y}[k]\}_{k \in \mathcal{P}}$. By applying LS estimation, interpolation, and domain transformation, we obtain a coarse angular-delay domain representation $\mathbf{C}_\mathrm{pilot}$. This tensor captures the sparse multi-path clusters, providing the primary skeleton for reconstruction. Finally, the DiT generates the target channel estimation $\hat{\mathbf{H}}$ via flow matching. Conditioned on both the structural skeleton $\mathbf{C}_\mathrm{pilot}$ and the environmental semantics $\mathbf{C}_\mathrm{env}$, 
the model learns to transport Gaussian noise to the clean channel distribution in a deterministic path. Below, we detail the specific data processing and network architecture.

\subsection{Multimodal Encoding and Fusion}
We fuse heterogeneous sensing data using a perceiver resampler and a transformer-based fusion module. By projecting modalities into a shared latent space, this compresses diverse inputs into compact tokens while preserving channel-relevant geometric and semantic features.

\subsubsection{Multimodal Encoding}
We process the three distinct modalities through specialized branches before fusion.

\textbf{LiDAR Branch}: We project raw point clouds into a multi-channel bird's eye view (BEV) encoding height and density. To handle the sparse, high-dimensional BEV map, we use a perceiver resampler \cite{Flamingo, MDT} with learnable query tokens that extract relevant features via cross-attention. Given BEV features $\mathbf{M}_\mathrm{lidar} \in \mathbb{R}^{H \times W \times C}$, the resampler outputs $\mathbf{Z}_\mathrm{lidar} \in \mathbb{R}^{N_\mathrm{lidar} \times D}$, where $N_\mathrm{lidar}$ is the token sequence length and $D$ is the embedding dimension.
This compresses LiDAR data into a compact representation by selectively gathering environment-relevant information.

\textbf{Camera Branch}: Four cameras provide $360^{\circ}$ coverage. We use a pre-trained ResNet-34 \cite{Res} to extract semantic features from each view, then apply a perceiver resampler to produce tokens $\mathbf{Z}_\mathrm{cam} \in \mathbb{R}^{N_\mathrm{cam} \times D}$. Learnable camera ID embeddings are injected to distinguish front, rear, left, and right views.

\textbf{Positioning Information Branch}: The relative position and velocity determine large-scale fading such as path loss and shadowing. For each entity, we construct a 4-dimensional feature vector $\mathbf{v}_{\mathrm{phy}} = [x, y, z, \log(d)]$, where $(x, y, z)$ are CAV-centric relative coordinates and $d$ is the Euclidean distance. We then project it via a multi-layer perceptron into tokens $\mathbf{Z}_{\mathrm{phy}} \in \mathbb{R}^{N_\mathrm{phy} \times D}$.

\subsubsection{Unified Transformer Fusion}
To model the complex interactions between different modalities, we employ a multimodal transformer encoder for deep fusion.
We first concatenate the tokens from all branches to form a heterogeneous sequence:
\begin{equation}
    \mathbf{Z}_\mathrm{mul} = [\mathbf{Z}_\mathrm{lidar}, \mathbf{Z}_\mathrm{cam}, \mathbf{Z}_\mathrm{phy}] \in \mathbb{R}^{(N_\mathrm{lidar} + N_\mathrm{cam} + N_\mathrm{phy}) \times D}.
\end{equation}
To retain modality-specific information, we add learnable modality embeddings to tokens belonging to each sensor type. Additionally, learnable positional embeddings are added to encode the sequence order.
The combined sequence is then processed by a multi-layer transformer encoder with self-attention. The output is the multimodal semantic condition $\mathbf{C}_\mathrm{env}$, which serves as the semantic prompt for the subsequent generative process.

% \subsubsection{Auxiliary Task Learning}
% To further encourage the Multimodal Conditioner to learn physically meaningful representations, we introduce an auxiliary classification head attached to the fusion output. This head is trained to predict high-level scene attributes (e.g., LOS/NLOS status or scenario ID) from the global average-pooled features. This multi-task learning strategy acts as a regularizer, ensuring the latent tokens capture valid environmental descriptors even when the gradient from the diffusion loss is noisy.

\subsection{Diffusion Transformer Architecture}
% 标题是否需要改一下
We design a DiT framework that integrates a dual-stream conditioning strategy: structural guidance from pilots and semantic guidance from the environment. By transforming channel data into the angle-delay domain and employing flow matching, the model learns a deterministic transport mapping for efficient high-fidelity reconstruction.

\subsubsection{Condition Formulation and Tokenization}
First, we derive the structural pilot condition from sparse observations $\{\mathbf{Y}[k]\}_{k \in \mathcal{P}}$. We perform LS estimation with linear interpolation to obtain a coarse estimate $\hat{\mathbf{H}}_{\mathrm{interp}}$. To exploit physical sparsity, we transform this estimate to the angle-delay domain via 2D discrete Fourier transforms and separate real-imaginary components, yielding the pilot tensor $\mathbf{C}_\mathrm{pilot} \in \mathbb{R}^{N_r \times N_t \times 2N_c}$.

Simultaneously, we prepare the input for the flow matching process. We apply the same domain transformation to the ground truth channel $\mathbf{H}$ to obtain the target tensor $\mathbf{H}_1$, which matches the dimensions of $\mathbf{C}_\mathrm{pilot}$. Adopting the conditional optimal transport formulation \cite{FM}, we sample a time step $t \sim \mathcal{U}[0,1]$ and linearly interpolate between the target data $\mathbf{H}_1$ and standard Gaussian noise $\mathbf{H}_0$. This yields the noisy state $\mathbf{H}_t = t \mathbf{H}_1 + (1-t) \mathbf{H}_0$, which corresponds to a straight-line probability path that minimizes transport cost and simplifies the learning objective.

Both $\mathbf{H}_t$ and $\mathbf{C}_\mathrm{pilot}$ are then processed via 3D patching. We partition the tensors into blocks of size $(p_\mathrm{ant}, p_\mathrm{ant}, p_\mathrm{freq})$, flatten them, and linearly project them to the DiT hidden dimension $D$. This results in the noisy channel tokens $\mathbf{E}_\mathrm{ch} \in \mathbb{R}^{L \times D}$ and the pilot condition tokens $\mathbf{E}_\mathrm{pilot} \in \mathbb{R}^{L \times D}$, where $L$ is the number of patches.

\subsubsection{Condition Injection Mechanism}
Once tokenized, the DiT integrates these inputs through distinct pathways to maximize their efficacy:

\textbf{Pilot Condition (Concat)}: The pilot tokens $\mathbf{E}_\mathrm{pilot}$ serve as a strong condition. They are explicitly concatenated with the noisy channel tokens $\mathbf{E}_\mathrm{ch}$ and augmented with learnable 3D positional embeddings $\mathbf{P}_\mathrm{pos}$. This combined sequence $\mathbf{E}_\mathrm{DiT} = [\mathbf{E}_\mathrm{ch}, \mathbf{E}_\mathrm{pilot}] + \mathbf{P}_\mathrm{pos}$ serves as the primary input to the backbone, directly providing the network with the coarse structure of the channel.

\textbf{Environmental Semantic Condition (Cross-Attention)}: The multimodal semantic condition $\mathbf{C}_\mathrm{env}$ serves as a specific context prompt. Unlike the pilot tokens, $\mathbf{C}_\mathrm{env}$ is injected into each DiT block via multi-head cross-attention layers. This allows the model to dynamically query environmental geometry and semantics to refine the details of the reconstruction.

Finally, the interpolation timestep $t$ is injected into the network via adaptive layer normalization, modulating the channel statistics at each layer to guide the denoising trajectory.

\subsubsection{Flow Matching Objective}
The model is trained to predict the velocity field $\mathbf{v}_t = \frac{\mathrm{d}\mathbf{H}_t}{\mathrm{d}t} = \mathbf{H}_1 - \mathbf{H}_0$ that drives the flow from noise to data\cite{FM}. We minimize the mean squared error:
\begin{equation}
    \mathcal{L}(\theta) = \mathbb{E}_{t \sim \mathcal{U}[0,1]} \left[ \| v_{\theta}(\mathbf{H}_t, t, \mathbf{E}_\mathrm{pilot}, \mathbf{C}_\mathrm{env}) - (\mathbf{H}_1 - \mathbf{H}_0) \|^2 \right],
\end{equation}
where the network $v_{\theta}$ is conditioned on both sparse pilot tokens $\mathbf{E}_\mathrm{pilot}$ and environmental semantics $\mathbf{C}_\mathrm{env}$. By learning this optimal transport path, we enforce straight generation trajectories to maximize sampling efficiency.

\subsubsection{Inference with Classifier-Free Guidance (CFG)}
We use CFG \cite{CFG} to control the influence of the environmental semantic condition on channel estimation. During training, we drop $\mathbf{C}_\mathrm{env}$ with probability $p_{\mathrm{drop}}=0.1$, replacing it with a learnable null embedding $\mathbf{C}_{\emptyset}$ to learn the unconditional prior.
During the inference stage, we combine conditional and unconditional velocity predictions:
\begin{equation}
\begin{split}
    &\tilde{v}_{\theta}(\mathbf{H}_t , t) = v_{\theta}(\mathbf{H}_t, t, \mathbf{E}_\mathrm{pilot}, \mathbf{C}_{\emptyset}) \\
    &+ w \cdot \left(v_{\theta}(\mathbf{H}_t, t, \mathbf{E}_\mathrm{pilot}, \mathbf{C}_\mathrm{env}) - v_{\theta}(\mathbf{H}_t, t, \mathbf{E}_\mathrm{pilot}, \mathbf{C}_{\emptyset})\right),
\end{split}
\end{equation}
where $w$ controls the strength of environmental guidance relative to pilot observations. A lower $w$ (e.g., $w < 1$) prioritizes learned priors and pilot inputs, thereby enhancing robustness against unreliable environmental semantics, particularly in OOD scenarios. 

Finally, leveraging the near-straight velocity field learned via flow matching, we recover the tokens of the target channel in a single Euler step:
\begin{equation}
    \hat{\mathbf{H}}_1 = \mathbf{H}_0 + \tilde{v}_{\theta}(\mathbf{H}_0, 0).
\end{equation}
The estimated tokens are then reshaped and transformed back to the frequency domain to obtain the final channel estimate $\hat{\mathbf{H}}$. This one-step generation significantly reduces latency compared to multi-step diffusion.

\section{Simulation Results}
\subsection{Experimental Setup}
We use the Multimodal-Wireless dataset~\cite{MultimodalWireless}, containing 16 scenarios with diverse urban layouts. To augment the training data, we employ a subcarrier splitting strategy, dividing the full bandwidth of each frame into $16$ non-overlapping sub-bands. 
% Leveraging the fact that small-scale fading statistics remain consistent across adjacent sub-bands, we treat each segment as an independent training instance. 
This approach effectively expands the dataset size by a factor of 16, yielding approximately $850,000$ samples. Each sample includes an MIMO-OFDM channel realization ($N_\mathrm{r}=4, N_\mathrm{t}=4, N_\mathrm{c}=64$), 64-beam LiDAR point clouds, 4-view RGB images, and positioning information.

We use 15 scenarios as the source domain, randomly split into training, validation, and in-distribution test sets with a ratio of 8:1:1. This results in approximately $640,000$ samples for training and $80,000$ samples for in-distribution testing. We reserve the ``Town07 single-lane road'' scenario ($48,000$ samples) for OOD testing. Unlike the urban training data, ``Town07 single-lane road'' features a rural environment with sparse buildings and dense foliage, verifying the generalization capability of the proposed model.

Our DiT backbone uses $N_\mathrm{depth}=6$ layers, $N_\mathrm{head}=6$ heads, and dimension $D=384$. The channel data is tokenized using 3D patches of size $(2, 2, 8)$. The model is trained using the AdamW optimizer with a batch size of 96, a weight decay factor of $0.05$. The learning rate follows a cosine annealing schedule, decaying from $2 \times 10^{-4}$ to $1 \times 10^{-6}$. To enhance robustness, we randomize the pilot spacing $S_\mathrm{p}$ within $\{2, 4, 8\}$ and sample the pilot SNR uniformly from $[-10, 30]$ dB during training.

The primary performance metric is the normalized mean-squared error (NMSE), defined as $\mathbb{E}\big\{\|\mathbf{H}-\hat{\mathbf{H}}\|_F^2\big\} / \mathbb{E}\big\{\|\mathbf{H}\|_F^2\big\}$. Additionally, we utilize cosine similarity to evaluate the structural alignment and beamforming direction accuracy, calculated as $\mathbb{E}\big\{ |\text{tr}(\mathbf{H}^H \hat{\mathbf{H}}) / \|\mathbf{H}\|_F \|\hat{\mathbf{H}}\|_F \big\}$, where $\|\cdot\|_F$ denotes the Frobenius norm.

% \subsection{Performance}
% \begin{figure}[!t]
%     \centering
%     \includegraphics[width=3.6in]{results_snr.pdf}
%     \caption{NMSE and cosine similarity performance versus SNR ($S_\mathrm{p}=8$).}
%     \label{fig:snr}
% \end{figure}
\begin{figure}[!t]
    \centering
    % \makebox 允许内部内容总宽度超过 \linewidth，但强制整体居中
    \makebox[\linewidth][c]{
        \subfigure[NMSE versus SNR.]{
            % 这里宽度可以大胆设为 0.52\linewidth 甚至更大
            \includegraphics[width=0.50\linewidth]{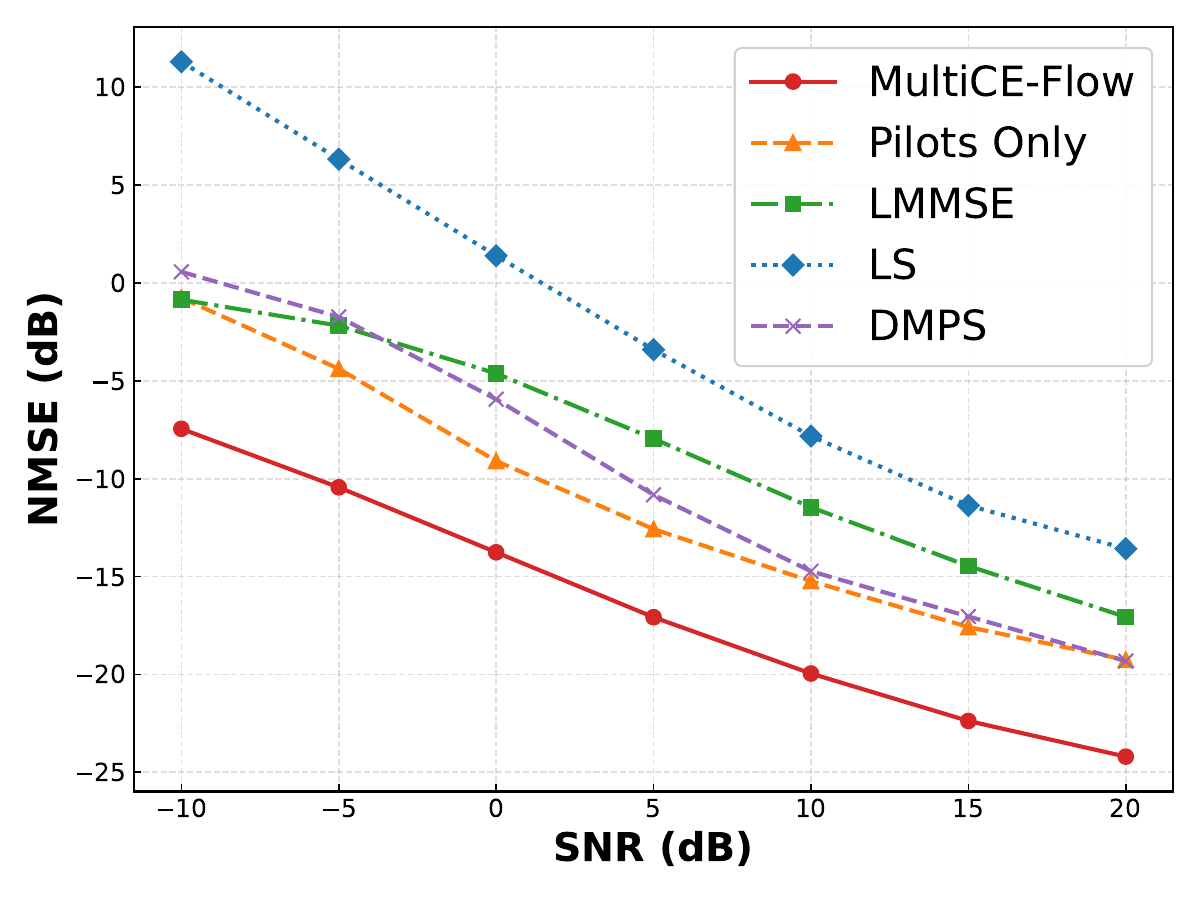}
            \label{fig:snr_nmse}
        }\hfill%
        \subfigure[Cosine similarity versus SNR.]{
            \includegraphics[width=0.50\linewidth]{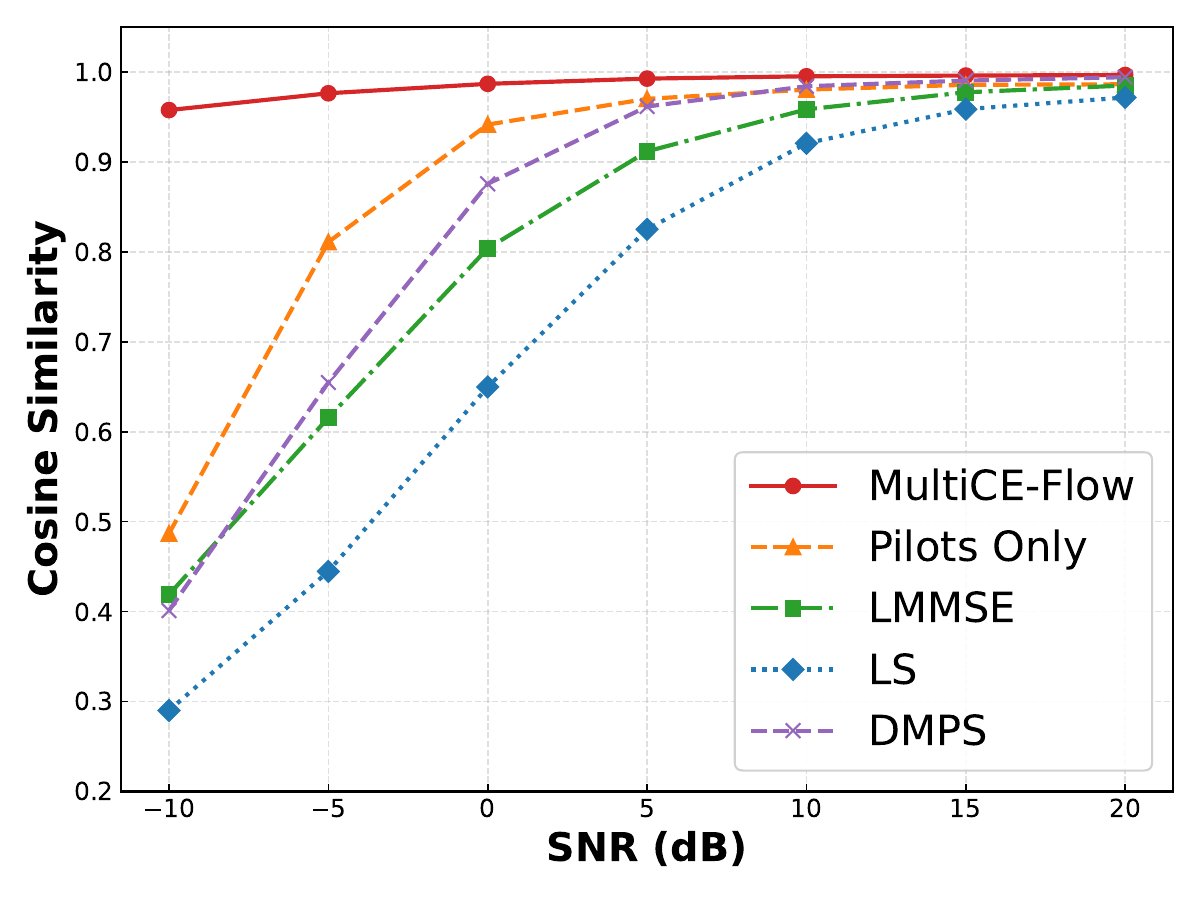}
            \label{fig:snr_cosine}
        }
    }
    \caption{Performance of MultiCE-Flow versus SNR ($S_\mathrm{p}=8$).}
    \label{fig:performance_snr}
\end{figure}

We compare our proposed MultiCE-Flow with the following baselines:
\textbf{LMMSE:} The linear estimator utilizing the full channel covariance matrix computed from the testing data.
\textbf{DMPS \cite{HGM}:} A diffusion-based estimator trained to learn the unconditional channel prior, employing posterior sampling for channel estimation. To ensure fair comparison, this baseline utilizes the same DiT backbone as our multimodal model.
\textbf{Pilots Only:} The MultiCE-Flow model operating without multimodal conditions (guidance scale $w=0$).

% \begin{figure}[!t]
%     \centering
%     \includegraphics[width=2in]{results_spacing.pdf}
%     \caption{Impact of pilot spacing on estimation accuracy.}
%     \label{fig:spacing}
% \end{figure}

We first evaluate the estimation accuracy under varying SNR conditions. As illustrated in Fig.~\ref{fig:performance_snr}, our multimodal method consistently outperforms all baselines across the entire SNR range.
The LS estimator suffers from severe degradation, particularly in the low SNR regime, due to its inherent inability to suppress noise.
While the LMMSE estimator improves upon LS by leveraging second-order channel statistics, it remains constrained by its linear modeling capabilities.
Notably, DMPS underperforms the Pilots Only method at low SNRs. This is because iterative posterior sampling is prone to instability under significant noise. In contrast, the Pilots Only approach, operating as a conditional generative model, effectively treats estimation as a robust denoising task guided by pilot tokens.
Our multimodal approach achieves the lowest NMSE and highest cosine similarity across the entire SNR range. This demonstrates that integrated environmental semantics provide critical informative guidance, effectively compensating for signal ambiguity when pilot observations are extremely sparse.

\begin{figure}[!t]
    \centering
    % 使用 \makebox 强制居中，允许内部元素总宽度超过 \linewidth
    \makebox[\linewidth][c]{
        % 左子图：宽度大胆突破 0.5，比如设为 0.53\linewidth
        \subfigure[Impact of pilot spacing.]{
            \includegraphics[width=0.485\linewidth]{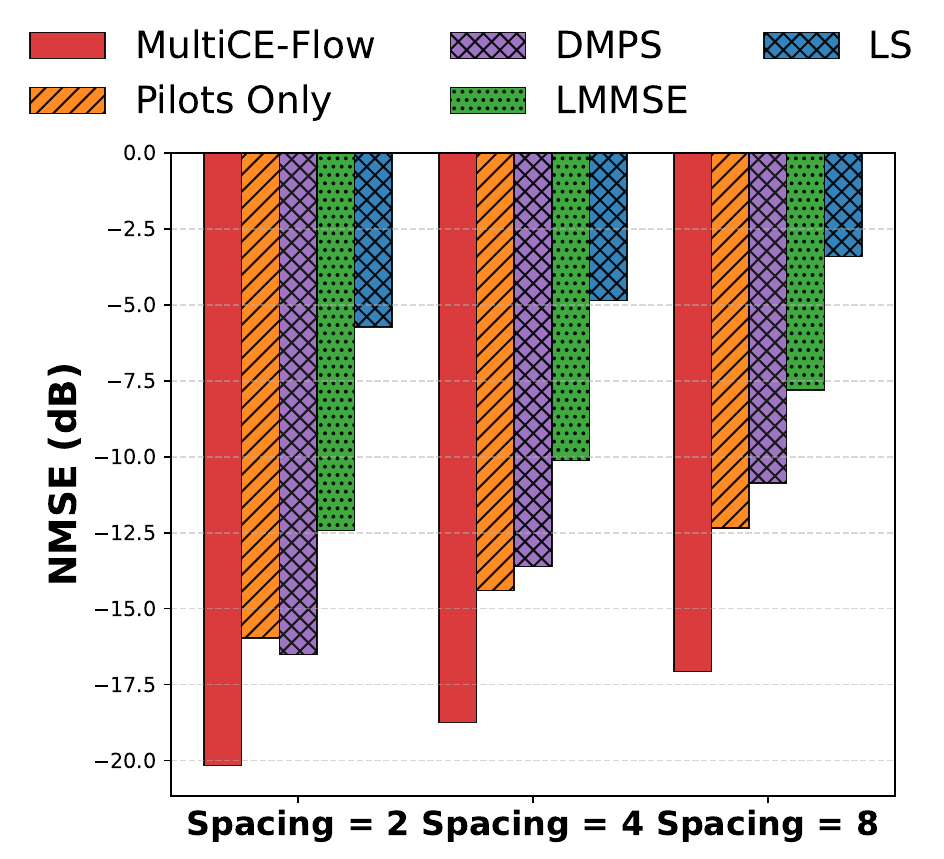}
            \label{fig:spacing}
        }\hspace{0.02\linewidth}% <--- 【关键】这里用 \hspace 手动指定一个小间距，并紧跟 % 消除回车换行
        % 右子图：同样设为 0.53\linewidth
        \subfigure[Ablation study ($S_\mathrm{p}=8$).]{
            \includegraphics[width=0.485\linewidth]{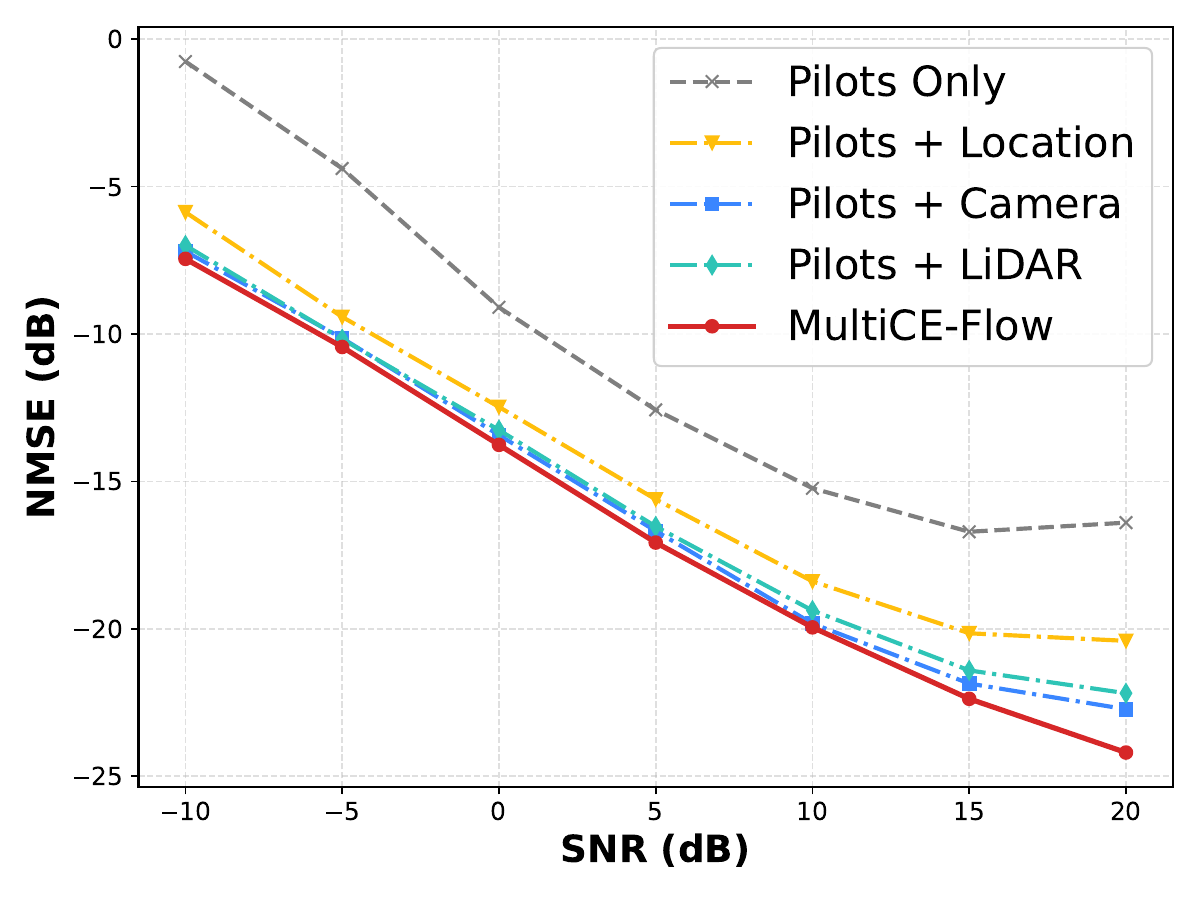}
            \label{fig:ablation}
        }
    }
    \caption{Performance analysis of MultiCE-Flow.}
    \label{fig:performance_analysis}
\end{figure}

Fig.~\ref{fig:spacing} evaluates the robustness of MultiCE-Flow under varying pilot spacing $S_\mathrm{p} \in \{2, 4, 8\}$. While all compared methods degrade as pilot sparsity increases, MultiCE-Flow consistently maintains a significant performance gain. Notably, even with extremely sparse pilots ($S_\mathrm{p}=8$), the proposed framework achieves an NMSE of -17 dB, outperforming the LMMSE estimator operating with dense pilots ($S_\mathrm{p}=2$). This demonstrates that environmental semantics effectively compensate for the insufficient channel observations in sparse-pilot scenarios.

The ablation study in Fig.~\ref{fig:ablation} quantifies the individual impact of each data modality. The Pilots Only baseline exhibits the poorest performance, reflecting the difficulty of reconstructing high-dimensional channels from sparse observations alone. Incorporating location data provides an approximately 5 dB gain at -10 dB SNR by leveraging large-scale fading correlations. Furthermore, incorporating visual modalities (LiDAR and Camera) yield substantial improvements by resolving scattering geometries. The performance of the full MultiCE-Flow framework confirms that fusing diverse environmental semantics is critical for robust channel estimation.

% \begin{figure}[!t]
%     \centering
%     \includegraphics[width=2in]{results_ablation.pdf}
%     \caption{Ablation study on different multimodal conditions (pilot spacing = 8).}
%     \label{fig:ablation}
% \end{figure}

% A critical challenge is generalization to unseen environments. We evaluate performance on the "Town07 single-lane road", a rural scenario never seen during training. As shown in Fig.~\ref{fig:ood}, while generative baselines outperform linear estimators like LS and LMMSE, they still face limitations. DMPS suffers from distribution shift, and Pilots Only lacks environmental context.
% In contrast, our multimodal method demonstrates superior OOD robustness. Note that we set the CFG scale to $w=0.5$ to conservatively balance the guidance from potentially shifted environmental cues with the robust prior conditioned on pilot observations. Our methods improves NMSE by over 3 dB compared to the Pilots Only method at low SNR. Crucially, it maintains high cosine similarity ($>0.86$) across all SNRs, ensuring precise beam alignment. This confirms that our model learns generalizable physical mappings from geometry to channel structure rather than simply memorizing training layouts.

\begin{figure}[!t]
    \centering
    % 用 \makebox 包装，允许内部总宽度突破 \linewidth 并强制整体居中
    \makebox[\linewidth][c]{
        \subfigure[NMSE versus SNR.]{
            % 宽度大胆放大，比如提升到 0.52 或 0.54
            \includegraphics[width=0.49\linewidth]{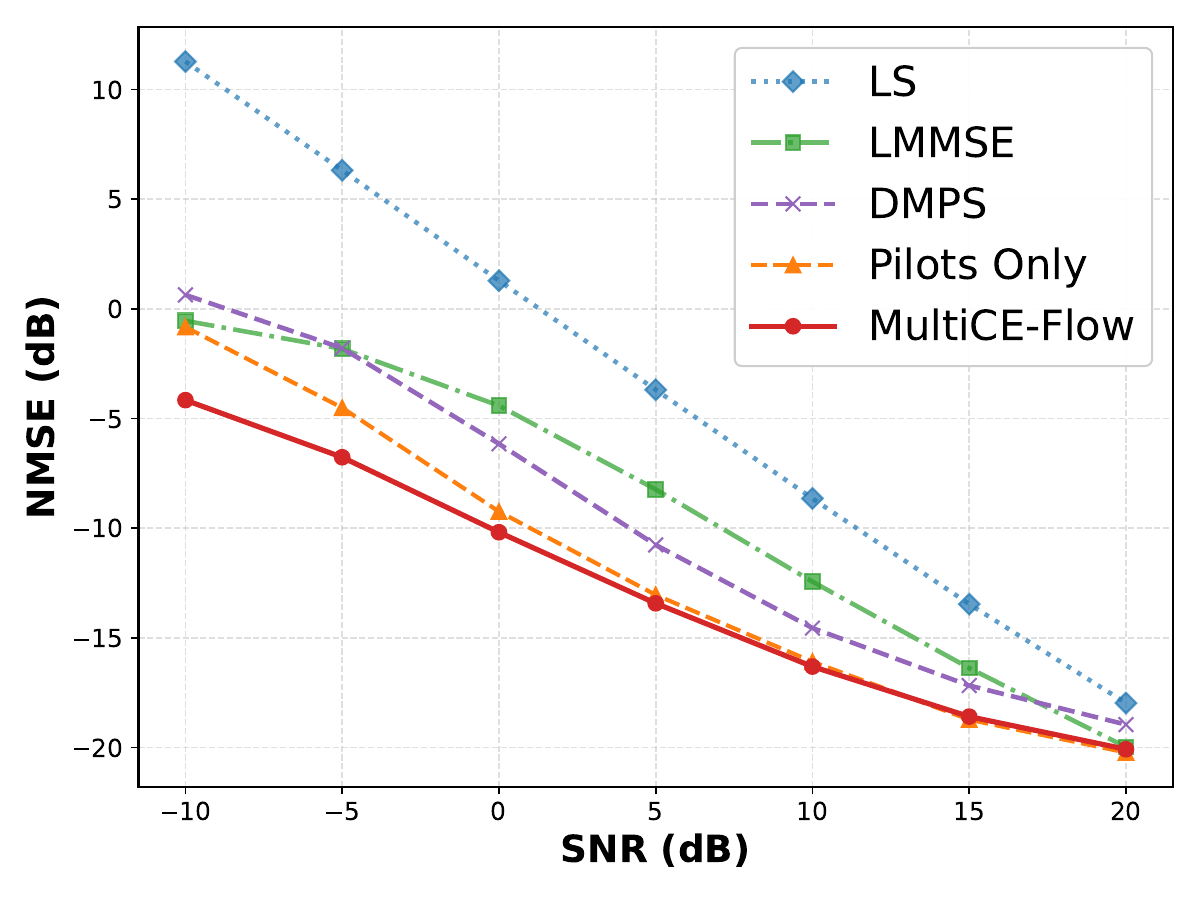}
            \label{fig:ood_nmse}
        }\hspace{0.02\linewidth}% <--- 【关键】手动设定微小间距，并用 % 吃掉换行符
        \subfigure[Cosine similarity versus SNR.]{
            % 宽度同步放大
            \includegraphics[width=0.49\linewidth]{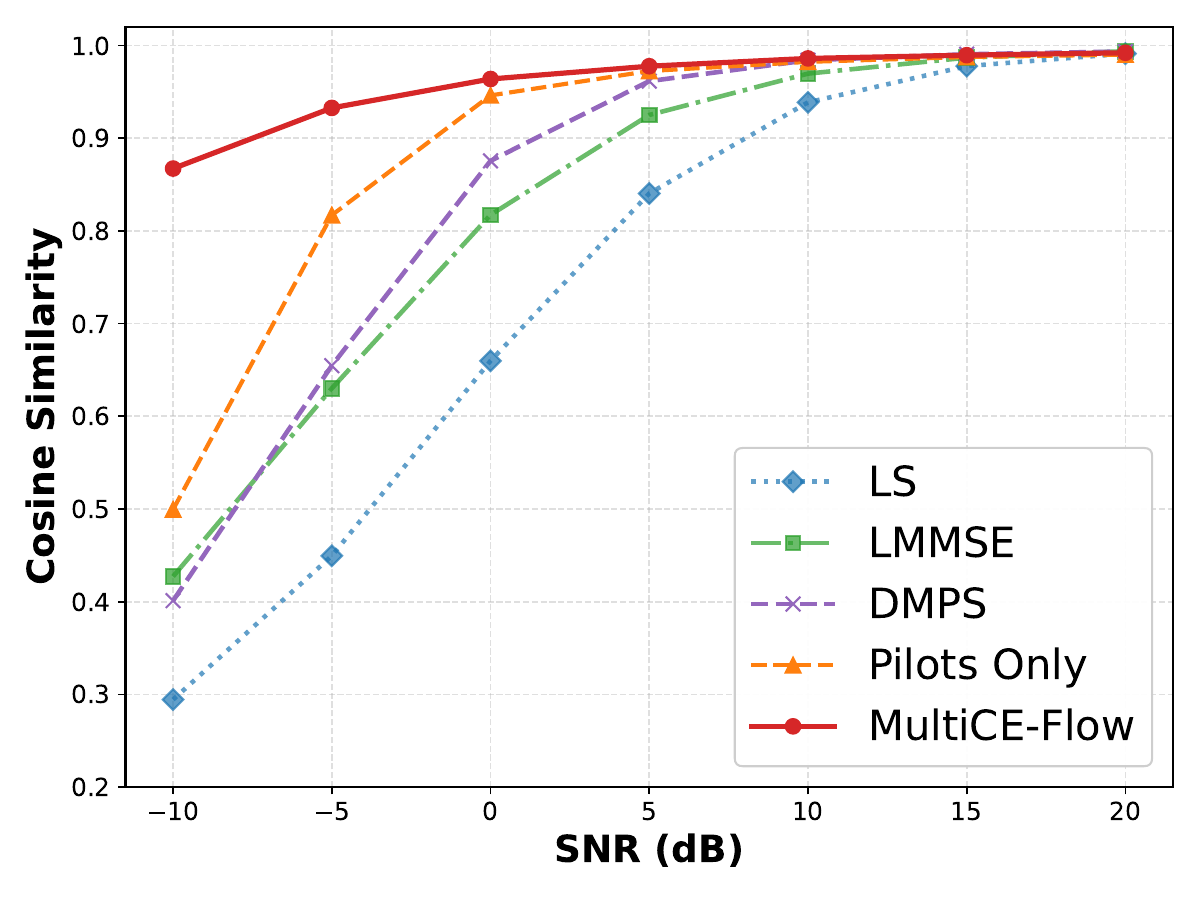}
            \label{fig:ood_cosine}
        }
    }
    \caption{Generalization performance of MultiCE-Flow ($S_\mathrm{p}=8$).}
    \label{fig:performance_ood}
\end{figure}

To evaluate generalization, we test on the unseen ``Town07 single-lane road'' scenario. As shown in Fig.~\ref{fig:performance_ood}, our multimodal approach demonstrates superior OOD robustness. By setting the CFG scale as $w=0.5$ to balance environmental guidance, our method improves NMSE by over 3 dB against the Pilots Only method at low SNR and maintains high cosine similarity ($>0.86$) across all SNRs. This confirms that environmental semantics learned during training effectively generalize to unseen scenarios.

Regarding computational complexity, the LMMSE estimator is lightweight but incurs a notable latency (2.9~ms) due to the heavy calculation of the $1024 \times 1024$ covariance matrix. The diffusion-based DMPS, using the same DiT backbone, suffers from prohibitive costs (69.4~GFLOPs, 5.8~ms) due to iterative sampling. In contrast, our MultiCE-Flow achieves superior efficiency: by leveraging single-step inference and sharing multimodal conditions for every 10 frames, it reduces the complexity to just 5.6~GFLOPs and 1.2~ms per sample, demonstrating superior real-time feasibility.

% \begin{figure}[!t]
%     \centering
%     \includegraphics[width=3.6in]{results_ood_generalization.pdf}
%     \caption{Generalization performance in an unseen OOD scenario ($S_\mathrm{p}=8$).}
%     \label{fig:ood}
% \end{figure}

\section{Conclusion}
This paper proposes MultiCE-Flow, a framework that reformulates channel estimation as a conditional generation task using flow matching and DiT. We introduce a dual-conditioning strategy that integrates sparse pilots as structural guidance and fused multimodal sensing data as semantic guidance. To support efficient inference, we employ flow matching with an optimal transport path to learn a deterministic transport map, simplifying the generation process into a straight trajectory. Experimental results show that MultiCE-Flow achieves superior reconstruction accuracy compared to traditional and generative baselines, particularly in challenging out-of-distribution environments. Moreover, the proposed method enables high-fidelity channel recovery with a single sampling step, satisfying the low-latency requirements of real-time communication systems.

% \nocite{*}
\bibliographystyle{IEEEtran}
% \bibliography{reference.bib}

\end{document}